\documentclass[10pt,twocolumn,letterpaper]{article}

\usepackage{cvpr}              

\usepackage{graphicx}
\usepackage{amsmath}
\usepackage{amssymb}
\usepackage{booktabs}
\usepackage{comment}
\usepackage{multirow}
\usepackage{mathtools}
\usepackage{epsfig}

\usepackage{multirow}
\usepackage[table,xcdraw,dvipsnames]{xcolor}

\usepackage[breaklinks,colorlinks,citecolor = ForestGreen]{hyperref}

\usepackage[capitalize]{cleveref}
\crefname{section}{Sec.}{Secs.}
\Crefname{section}{Section}{Sections}
\Crefname{table}{Table}{Tables}
\crefname{table}{Tab.}{Tabs.}

\def\cvprPaperID{2} 
\def\confName{CVPRW}
\def\confYear{2023}

\title{Flexible-Modal Face Anti-Spoofing: A Benchmark}

\begin{document}

\author{Zitong Yu\textsuperscript{1}, Ajian Liu\textsuperscript{2}, Chenxu Zhao\textsuperscript{3}, Kevin H. M. Cheng\textsuperscript{4}, Xu Cheng\textsuperscript{5}, Guoying Zhao\textsuperscript{4}\\
\textsuperscript{1}Great Bay University, China  \\ \textsuperscript{2}Institute of Automation, Chinese Academy of Sciences, China  \\
\textsuperscript{3}SailYond Technology, China  \\
\textsuperscript{4}University of Oulu, Finland\\
\textsuperscript{5}Nanjing University of
Information Science and Technology, China }

\maketitle

\begin{abstract}

Face anti-spoofing (FAS) plays a vital role in securing face recognition systems from presentation attacks. Benefitted from the maturing camera sensors, single-modal (\textit{RGB}) and multi-modal (e.g., \textit{RGB+Depth}) FAS has been applied in various scenarios with different configurations of sensors/modalities. Existing single- and multi-modal FAS methods usually separately train and deploy models for each possible modality scenario, which might be redundant and inefficient. Can we train a unified model, and flexibly deploy it under various modality scenarios? In this paper, we establish the first flexible-modal FAS benchmark with the principle `train one for all'. To be specific, with trained multi-modal (\textit{RGB+Depth+IR}) FAS models, both intra- and cross-dataset testings are conducted on four flexible-modal sub-protocols (\textit{RGB}, \textit{RGB+Depth}, \textit{RGB+IR}, and \textit{RGB+Depth+IR}). We also investigate prevalent deep models and feature fusion strategies for flexible-modal FAS. We hope this new benchmark will facilitate the future research of the multi-modal FAS. The protocols and codes are available at \href{https://github.com/ZitongYu/Flex-Modal-FAS}{https://github.com/ZitongYu/Flex-Modal-FAS}.

\end{abstract}


\section{Introduction}

\thispagestyle{empty}

\begin{figure}[t]
\centering
\includegraphics[scale=0.35]{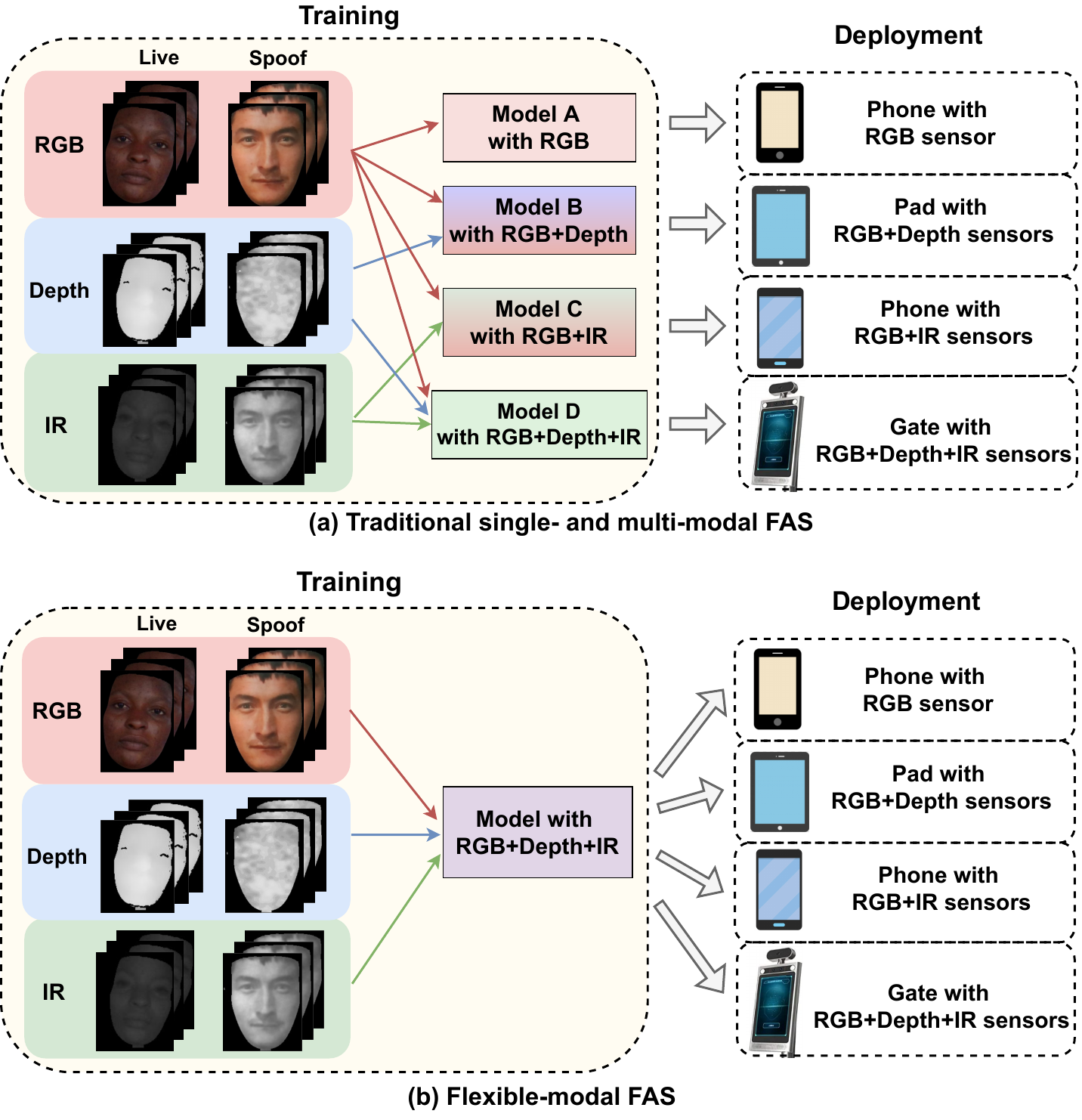}
 \vspace{-1.5em}
  \caption{\small{
  The training and deployment frameworks of the (a) traditional single- and multi-modal FAS; and (b) flexible-modal FAS. The former one aims at separately train and deploy powerful models for each possible modality scenario, while the latter one focuses on training a unified model for all real-world modality scenarios.}
  }
 
\label{fig:Figure1}

\end{figure}

Face recognition has been widely used in many interactive artificial intelligence systems for its convenience (e.g., access control and face payment). However, vulnerability to presentation attacks (e.g., print, video replay, and 3D masks) curtails its reliable deployment. For the reliable use of face recognition systems, face anti-spoofing (FAS) methods~\cite{yu2021deep,yu2021revisiting} are important to detect such presentation attacks (PAs). 


In recent years, plenty of hand-crafted feature based~\cite{boulkenafet2015face,Boulkenafet2017Face,yu2021transrppg} and deep learning based~\cite{liu2018Learning,yu2020searching,wang2020deep,qin2019learning,yu2020fas,yu2020face,yu2021dual,qin2021meta,cai2020drl,cai2022learning,liu2022contrastive} methods have been proposed for \textit{RGB}-based single-modal FAS. On one hand, some hand-crafted descriptors with facial color texture~\cite{boulkenafet2015face} and physiological signals~\cite{yu2021transrppg} feature representation are designed based on crucial live/spoof clues (e.g., moir$\rm\acute{e}$ pattern, noise artifacts, and bio-signal liveness), thus are robust for live/spoof discrimination. On the other hand, deep convolutional neural networks (CNN)~\cite{he2016deep} and vision transformer (ViT)~\cite{dosovitskiy2020image,george2021effectiveness,yu2021transrppg} become mainstream in FAS due to their strong semantic representation capacities to distinguish the bonafide from PAs. 

\begin{figure*}[t]
\centering
\includegraphics[scale=0.5]{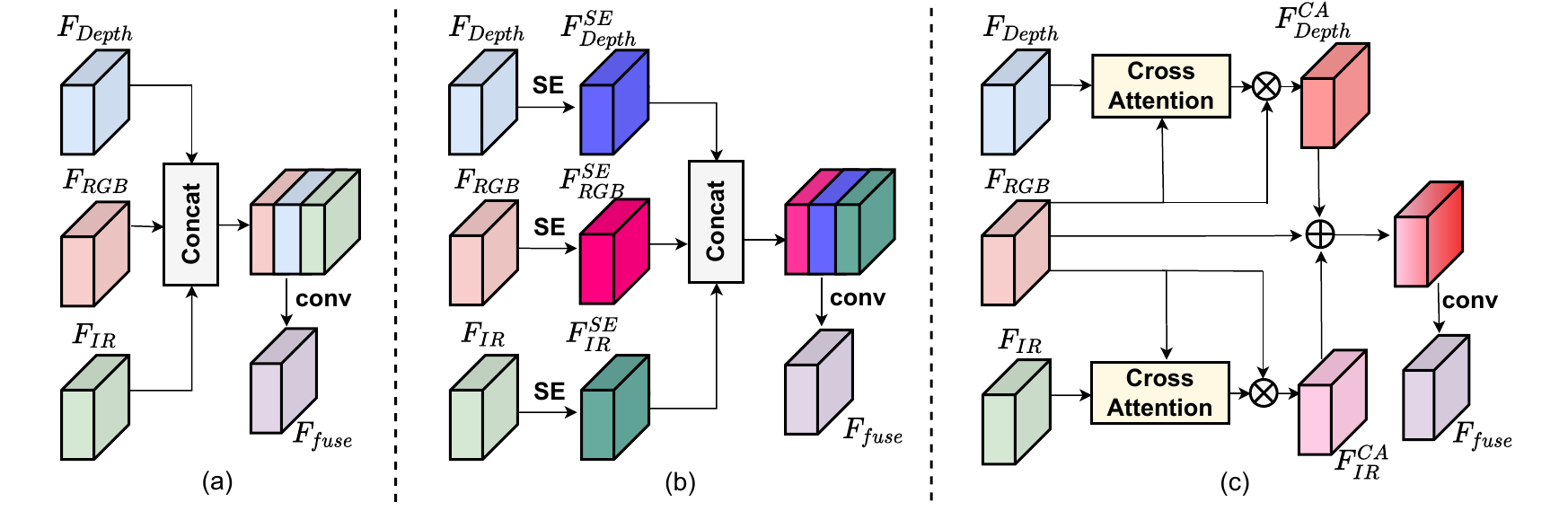}
\vspace{-0.8em}
  \caption{\small{
  Feature fusion modules. (a) Direct concatenation~\cite{yu2020multi}. (b) Squeeze-and-excitation (SE) fusion~\cite{casiasurf}. (c) Cross-attention fusion.}
  }
 \vspace{-0.5em}
\label{fig:network}

\end{figure*}

With the development of hardware manufacture and integration technology, multi-modal FAS systems with acceptable costs are increasingly used in real-world applications. Meanwhile, a few large-scale multi-modal FAS datasets~\cite{casiasurf,liu2021casia,george2019biometric} as well as multi-modal deep learning based FAS methods~\cite{yu2020multi,shen2019facebagnet,george2021cross,liu2021face,wang2019multi} are proposed. In terms of multi-modal FAS datasets, CASIA-SURF~\cite{casiasurf} and CeFA~\cite{liu2021casia} are with three modalities (\textit{RGB}, \textit{Depth}, and \textit{Infra-red (IR)}) while WMCA~\cite{george2019biometric} is with four modalities (\textit{RGB}, \textit{Depth}, \textit{IR}, and \textit{Thermal}). To learn intrinsic live/spoof features from multiple modalities, feature-level fusion strategies~\cite{yu2020multi,shen2019facebagnet,liuma2022,yu2022benchmarking,yu2023rethinking} are used. To better leverage contextual modality information and eliminate redundancy, spatial~\cite{wang2019multi} and channel~\cite{wang2019multi,casiasurf} attention is applied in multi-modal fusion.

Existing single- and multi-modal FAS methods usually separately train and deploy models for each possible modality scenario (see Fig.~\ref{fig:Figure1}(a)), which might be redundant and inefficient. A few natural questions occur: Can we train a unified model, and flexibly deploy it under various modality scenarios? How about the performance and efficiency gaps among separate and unified single- and multi-modal models? To explore the questions above, we establish the first flexible-modal FAS benchmark with the principle `train one for all', focusing on training a unified model for multiple real-world modality scenarios (see Fig.~\ref{fig:Figure1}(b)). Our contributions include:

\begin{itemize}
\setlength\itemsep{-0.1em}

    \item We establish the first flexible-modal FAS benchmark with both intra- and cross-dataset testings under four evaluation modality scenarios (\textit{RGB}, \textit{RGB+Depth}, \textit{RGB+IR}, and \textit{RGB+Depth+IR}).

    \item We propose an elegant cross-attention fusion module to efficiently mine cross-modal clues for flexible-modal deployment. The  proposed cross-Attention module significantly benefits the ViT~\cite{dosovitskiy2020image} in both flexible intra- and cross-testings.

    \item We also investigate prevalent deep models (CDCN~\cite{yu2020searching}, ResNet~\cite{he2016deep}, ViT~\cite{dosovitskiy2020image}) and feature fusion strategies for flexible-modal FAS. We find that the modality dropout strategy~\cite{shen2019facebagnet} works well in flex-modal intra-testings but poorly in flex-modal cross-testings.

\end{itemize}

\section{Multi-Modal Fusion Baselines}
\label{sec:method}
For the \textit{RGB}-based single-modal FAS task, given a face input $X_{RGB}$, the corresponding deep features/descriptors $F_{RGB}$ could be extracted. Then a prediction head $h$ is cascaded for binary live/spoof classification. For the \textit{RGB+Depth+IR} multi-modal FAS task, independent-modality features $F_{RGB}$, $F_{Depth}$, and $F_{IR}$ could be captured from face inputs $X_{RGB}$, $X_{Depth}$, and $X_{IR}$, respectively. All these features will be fused to form $F_{fuse}$ first, and then forward the prediction head $h$. In this paper, we focus on feature-level fusion strategy but there are also e.g., decision-level fusion (a late fusion strategy) for multi-modal scenarios. Here we discuss about three feature-level fusion modules under the scenario with \textit{RGB+Depth+IR}, and it is easily extended to scenarios with less or more modalities.

 \vspace{0.5em}
\noindent\textbf{Direct concatenation fusion.} Despite coarse alignment in the spatial domain, the features ($F_{RGB}$, $F_{Depth}$, and $F_{IR}$) in the channel domain have heterogeneous representation. As illustrated in Fig.~\ref{fig:network}(a), one classical solution is to concatenate these three features in the channel domain first~\cite{yu2020multi}, and then aggregate the multi-modal heterogeneous features with a lightweight fusion operator (e.g., convolution). The direct concatenation fusion can be formulated as    
\vspace{-0.3em}
\begin{equation} 
F_{fuse} = \mathrm{ReLU(BN(Conv(Concat}(F_{RGB},F_{Depth},F_{IR}))).
\vspace{-0.3em}
\end{equation}

 \vspace{0.5em}
\noindent\textbf{Squeeze-and-excitation fusion.}
To alleviate the feature misalignment among modalities, squeeze-and-excitation (SE) module~\cite{hu2018squeeze,casiasurf} is utilized in each independent modality branch first. With the channel-wise self-calibration via SE module, the refined features ($F^{SE}_{RGB}$, $F^{SE}_{Depth}$, and $F^{SE}_{IR}$) are then concatenated and aggregated. The framework of SE fusion is shown in Fig.~\ref{fig:network}(b). The SE fusion can be formulated as (where $\sigma$ denotes the Sigmoid function):    
\vspace{-0.3em}
\begin{equation}
\begin{split}
&F^{SE}_{RGB} = F_{RGB}\cdot\sigma \mathrm{(FC(ReLU(FC(AvgPool}(F_{RGB}))))),\\
&F^{SE}_{Depth} = F_{Depth}\cdot\sigma \mathrm{(FC(ReLU(FC(AvgPool}(F_{Depth}))))),\\
&F^{SE}_{IR} = F_{IR}\cdot\sigma \mathrm{(FC(ReLU(FC(AvgPool}(F_{IR}))))),\\
&F_{fuse} = \mathrm{ReLU(BN(Conv(Concat}(F^{SE}_{RGB},F^{SE}_{Depth},F^{SE}_{IR}))),
\end{split}
\vspace{-1.3em}
\end{equation}

 \vspace{0.6em}
\noindent\textbf{Cross-attention fusion.} Besides fusion via multi-modal feature concatenation, we also explore the feature addition in the homogeneous space. To this end, we calculate the relationship maps between $F_{RGB}$ and $F_{Depth}$/$F_{IR}$ via cross-attention (CA), and then the normalized modality-interacted maps are multiplied by $F_{RGB}$ to form cross-attentioned features $F^{CA}_{Depth}$ and $F^{CA}_{IR}$. Finally, original RGB feature and cross-attentioned features are added and aggregated with an extra convolution. The framework of CA fusion is shown in Fig.~\ref{fig:network}(c). The CA fusion can be formulated as   
\vspace{-0.3em}
\begin{equation}
\begin{split}
&\bar{F}^{CA}_{Depth}=\mathrm{Softmax}(\bar{F}_{Depth}(\bar{F}_{RGB})^{T})\bar{F}_{RGB},\\
&\bar{F}^{CA}_{IR}=\mathrm{Softmax}(\bar{F}_{IR}(\bar{F}_{RGB})^{T})\bar{F}_{RGB},\\
&F_{fuse} = \mathrm{ReLU(BN(Conv}(F_{RGB}+F^{CA}_{Depth}+F^{CA}_{IR})),
\end{split}
\vspace{-1.3em}
\end{equation}
where $F$ and $\bar{F}$ denote the spatial features and vectorized features, respectively.   

 \captionsetup[table]{labelformat=empty,labelsep = none}

\section{Flexible-Modal FAS Benchmark}
\label{sec:experiment}
In this section, we introduce the flexible-modal FAS benchmark in terms of datasets, modality-aware protocols, and evaluation metrics. Statistical description is shown in Table~\ref{tab:experiment1}. 

 \vspace{0.2em}
\noindent\textbf{Datasets.} \quad Three large-scale multi-modal datasets are used in the flexible-modal FAS benchmark. \textbf{CASIA-SURF}~\cite{casiasurf} consists of 1000 subjects with 21000 videos (7000 for RGB, Depth, and IR modality, respectively). There are two kinds of presentation attack instrument (PAI), i.e., print and cut print attacks, in CASIA-SURF. \textbf{CeFA}~\cite{liu2021casia} is a cross-ethnicity FAS dataset, covering three ethnicities (Africa,  East Asia, and Central Asia), three modalities (RGB, Depth, and IR), 1607 subjects with 7846 videos for each modality. In terms of PAIs, it consists of print, replay, 3D print and 3D silica gel mask attacks. \textbf{WMCA}~\cite{george2019biometric} consists of 1941 short video recordings of both bonafide and PAs from 72 different identities. Each video is recorded from several spectrum channels including RGB, depth, IR, and thermal. In addition, there are seven PAIs (i.e., glasses, fake head, print, replay, rigid mask, flexible mask, and paper mask) in WMCA.

 \vspace{0.2em}
\noindent\textbf{Protocols.}  Towards the principle `train one for all', four flexible-modal protocols are established. Specifically, after trained on CASIA-SURF and CeFA with \textit{RGB+Depth+IR} modalities, the unified multi-modal model is evaluated on both intra (CASIA-SURF and CeFA) and cross (WMCA) datasets under \textit{RGB}-based (Protocol 1) single-modal, \textit{RGB+ Depth}-based  (Protocol 2) or \textit{RGB+IR}-based  (Protocol 3) bi-modal, and \textit{RGB+Depth+IR}-based  (Protocol 4) tri-modal scenarios. We also compare it with traditional separate training framework in terms of performance and efficiency.

 \vspace{0.2em}
\noindent\textbf{Evaluation metrics.}\quad  For all experiments, ACER~\cite{ACER} and True Positive Rate (TPR)@False Positive Rate (FPR) are used as evaluation metrics. ACER calculates the mean of Attack Presentation Classification Error Rate (APCER) and Bona Fide Presentation Classification Error Rate (BPCER). The thresholds are determined by the \textit{Equal Error Rate (EER) threshold} on validation set and set as \textit{0.5} for intra- and cross-dataset testings, respectively. Besides, TPR@FPR=0.1\% and TPR@FPR=1\% are also utilized for fair comparison.

\section{Experiment}
\label{sec:experiment}

\begin{table}[]
\centering
\caption{\textbf{Table 1:} \small{Statistics of the flexible-modal FAS benchmark. `C.-S.' and `D' are short for `CASIA-SURF' and `Depth', respectively. `\#Video' indicates the video numbers of each modality.}}
\vspace{-0.8em}
\resizebox{0.48\textwidth}{!}{\begin{tabular}{cccccccc|}
\cline{4-8}
\multicolumn{3}{c|}{}                                                                     & \multicolumn{4}{c|}{\cellcolor[HTML]{EFEFEF}Intra-dataset}                                                                              & \cellcolor[HTML]{EFEFEF}Cross-dataset \\ \hline
\multicolumn{1}{|c|}{Partition}  & \multicolumn{2}{c|}{Training}                          & \multicolumn{2}{c|}{Validation}                        & \multicolumn{2}{c|}{Testing}                           & Testing       \\ \hline
\multicolumn{1}{|c|}{Dataset}    & \multicolumn{1}{c|}{C.-S.} & \multicolumn{1}{c|}{CeFA} & \multicolumn{1}{c|}{C.-S.} & \multicolumn{1}{c|}{CeFA} & \multicolumn{1}{c|}{C.-S.} & \multicolumn{1}{c|}{CeFA} & WMCA          \\ \hline
\multicolumn{1}{|c|}{\#Subject}  & \multicolumn{1}{c|}{300}   & \multicolumn{1}{c|}{600}  & \multicolumn{1}{c|}{100}   & \multicolumn{1}{c|}{300}  & \multicolumn{1}{c|}{600}   & \multicolumn{1}{c|}{699}  & 72            \\ \hline
\multicolumn{1}{|c|}{\#Video}    & \multicolumn{1}{c|}{2100}  & \multicolumn{1}{c|}{2400} & \multicolumn{1}{c|}{700}   & \multicolumn{1}{c|}{1200} & \multicolumn{1}{c|}{4200}  & \multicolumn{1}{c|}{4246} & 1679          \\ \hline
\multicolumn{1}{|c|}{\#PAI}  & \multicolumn{1}{c|}{2}     & \multicolumn{1}{c|}{2}    & \multicolumn{1}{c|}{2}     & \multicolumn{1}{c|}{2}    & \multicolumn{1}{c|}{2}     & \multicolumn{1}{c|}{4}    & 7             \\ \hline
\multicolumn{1}{c|}{}            & \multicolumn{7}{c|}{\cellcolor[HTML]{EFEFEF}Modality}                                                                                                                                                            \\ \hline
\multicolumn{1}{|c|}{Protocol 1} & \multicolumn{2}{c|}{RGB+D+IR}                          & \multicolumn{2}{c|}{RGB}                               & \multicolumn{2}{c|}{RGB}                               & RGB           \\ \hline
\multicolumn{1}{|c|}{Protocol 2} & \multicolumn{2}{c|}{RGB+D+IR}                          & \multicolumn{2}{c|}{RGB+D}                             & \multicolumn{2}{c|}{RGB+D}                             & RGB+D         \\ \hline
\multicolumn{1}{|c|}{Protocol 3} & \multicolumn{2}{c|}{RGB+D+IR}                          & \multicolumn{2}{c|}{RGB+IR}                            & \multicolumn{2}{c|}{RGB+IR}                            & RGB+IR        \\ \hline
\multicolumn{1}{|c|}{Protocol 4} & \multicolumn{2}{c|}{RGB+D+IR}                          & \multicolumn{2}{c|}{RGB+D+IR}                          & \multicolumn{2}{c|}{RGB+D+IR}                          & RGB+D+IR      \\ \hline
\end{tabular}}
\label{tab:protocol}
\end{table}

\begin{table*}[t]
  \vspace{-0.8em}
\centering
\caption{\textbf{Table 2:} \small{Results of intra-dataset testings on CASIA-SURF and CeFA datasets with `Separate' and `Unified' settings.}}
 \vspace{-0.8em}
\resizebox{1.0\textwidth}{!}{\begin{tabular}{|cl|cc|cc|cc|cc|}
\hline
\multicolumn{2}{|c|}{}                                                                                & \multicolumn{2}{c|}{Protocol 1}                                                             & \multicolumn{2}{c|}{Protocol 2}                                                             & \multicolumn{2}{c|}{Protocol 3}                                                             & \multicolumn{2}{c|}{Protocol 4}                                                             \\ \cline{3-10} 
\multicolumn{2}{|c|}{\multirow{-2}{*}{Method}}                                                        & \multicolumn{1}{l|}{ACER(\%) $\downarrow$}                      & \multicolumn{1}{l|}{TPR(\%)@FPR=0.1\% $\uparrow$} & \multicolumn{1}{l|}{ACER(\%) $\downarrow$}                      & \multicolumn{1}{l|}{TPR(\%)@FPR=0.1\% $\uparrow$} & \multicolumn{1}{l|}{ACER(\%) $\downarrow$}                      & \multicolumn{1}{l|}{TPR(\%)@FPR=0.1\% $\uparrow$} & \multicolumn{1}{l|}{ACER(\%) $\downarrow$}                      & \multicolumn{1}{l|}{TPR(\%)@FPR=0.1\% $\uparrow$} \\ \hline
\multicolumn{1}{|c|}{}                            & CDCN \cite{yu2020searching}                                              & \multicolumn{1}{c|}{32.49}                         & 4.7                                    & \multicolumn{1}{c|}{6.22}                          & 47.4                                   & \multicolumn{1}{c|}{43.98}                         & 0.93                                   & \multicolumn{1}{c|}{6.46}                          & 55.43                                  \\ 
\multicolumn{1}{|c|}{}                            & ResNet50 \cite{he2016deep}                                          & \multicolumn{1}{c|}{10.41}                         & 45.93                                  & \multicolumn{1}{c|}{1.7}                           & 88.23                                  & \multicolumn{1}{c|}{41.26}                         & 3.13                                   & \multicolumn{1}{c|}{3.07}                          & 62.33                                  \\ 
\multicolumn{1}{|c|}{\multirow{-3}{*}{Separate}} & ViT-Base \cite{dosovitskiy2020image}                                          & \multicolumn{1}{c|}{10.81}                         & 31.33                                  & \multicolumn{1}{c|}{1.44}                          & 91.27                                  & \multicolumn{1}{c|}{26.34}                         & 6.5                                    & \multicolumn{1}{c|}{3.82}                          & 80.87                                  \\ \hline
\multicolumn{1}{|c|}{}                            & CDCN                                              & \multicolumn{1}{c|}{32.69}                         & 2.43                                   & \multicolumn{1}{c|}{5.08}                          & 68.17                                  & \multicolumn{1}{c|}{41.32}                         & 0.6                                    & \multicolumn{1}{c|}{6.46}                          & 55.43                                  \\ 
\multicolumn{1}{|c|}{}                            & \cellcolor[HTML]{EFEFEF}CDCN w/ DropModal         & \multicolumn{1}{c|}{\cellcolor[HTML]{EFEFEF}36.49} & \cellcolor[HTML]{EFEFEF}1.47           & \multicolumn{1}{c|}{\cellcolor[HTML]{EFEFEF}7.91}  & \cellcolor[HTML]{EFEFEF}40.83          & \multicolumn{1}{c|}{\cellcolor[HTML]{EFEFEF}35.89} & \cellcolor[HTML]{EFEFEF}0.97           & \multicolumn{1}{c|}{\cellcolor[HTML]{EFEFEF}11.36} & \cellcolor[HTML]{EFEFEF}30.1           \\ 
\multicolumn{1}{|c|}{}                            & CDCN\_SE                                          & \multicolumn{1}{c|}{35.13}                         & 1.03                                   & \multicolumn{1}{c|}{4.6}                           & 61                                     & \multicolumn{1}{c|}{37.09}                         & 0.3                                    & \multicolumn{1}{c|}{8.81}                          & 46.8                                   \\ 
\multicolumn{1}{|c|}{}                            & \cellcolor[HTML]{EFEFEF}CDCN\_SE w/ DropModal     & \multicolumn{1}{c|}{\cellcolor[HTML]{EFEFEF}46.5}  & \cellcolor[HTML]{EFEFEF}0.3            & \multicolumn{1}{c|}{\cellcolor[HTML]{EFEFEF}25.88} & \cellcolor[HTML]{EFEFEF}19.6           & \multicolumn{1}{c|}{\cellcolor[HTML]{EFEFEF}46.8}  & \cellcolor[HTML]{EFEFEF}0.23           & \multicolumn{1}{c|}{\cellcolor[HTML]{EFEFEF}31.15} & \cellcolor[HTML]{EFEFEF}16.4           \\ 
\multicolumn{1}{|c|}{}                            & CDCN\_CA                                          & \multicolumn{1}{c|}{34.99}                         & 1.93                                   & \multicolumn{1}{c|}{31.55}                         & 2.37                                   & \multicolumn{1}{c|}{35.18}                         & 1.83                                   & \multicolumn{1}{c|}{34.33}                         & 2.2                                    \\ 
\multicolumn{1}{|c|}{}                            & \cellcolor[HTML]{EFEFEF}CDCN\_CA w/ DropModal     & \multicolumn{1}{c|}{\cellcolor[HTML]{EFEFEF}40.58} & \cellcolor[HTML]{EFEFEF}1.23           & \multicolumn{1}{c|}{\cellcolor[HTML]{EFEFEF}38.9}  & \cellcolor[HTML]{EFEFEF}1.43           & \multicolumn{1}{c|}{\cellcolor[HTML]{EFEFEF}39.38} & \cellcolor[HTML]{EFEFEF}1.2            & \multicolumn{1}{c|}{\cellcolor[HTML]{EFEFEF}40.94} & \cellcolor[HTML]{EFEFEF}1.2            \\ \cline{2-10} 
\multicolumn{1}{|c|}{}                            & ResNet50                                          & \multicolumn{1}{c|}{27.03}                         & 15.37                                  & \multicolumn{1}{c|}{2.67}                          & 80.93                                  & \multicolumn{1}{c|}{34.17}                         & 2.63                                   & \multicolumn{1}{c|}{3.07}                          & 62.33                                   \\ 
\multicolumn{1}{|c|}{}                            & \cellcolor[HTML]{EFEFEF}ResNet50 w/ DropModal     & \multicolumn{1}{c|}{\cellcolor[HTML]{EFEFEF}14.24} & \cellcolor[HTML]{EFEFEF}23.23          & \multicolumn{1}{c|}{\cellcolor[HTML]{EFEFEF}9.32}  & \cellcolor[HTML]{EFEFEF}61.5           & \multicolumn{1}{c|}{\cellcolor[HTML]{EFEFEF}18.18} & \cellcolor[HTML]{EFEFEF}22.77          & \multicolumn{1}{c|}{\cellcolor[HTML]{EFEFEF}8.1}   & \cellcolor[HTML]{EFEFEF}56.27          \\ 
\multicolumn{1}{|c|}{}                            & ResNet50\_SE                                      & \multicolumn{1}{c|}{20.59}                         & 6.87                                   & \multicolumn{1}{c|}{2.05}                          & 78.37                                  & \multicolumn{1}{c|}{27.67}                         & 2.3                                    & \multicolumn{1}{c|}{2.48}                          & 55.3                                   \\ 
\multicolumn{1}{|c|}{}                            & \cellcolor[HTML]{EFEFEF}ResNet50\_SE w/ DropModal     & \multicolumn{1}{c|}{\cellcolor[HTML]{EFEFEF}14.57} & \cellcolor[HTML]{EFEFEF}32.6           & \multicolumn{1}{c|}{\cellcolor[HTML]{EFEFEF}3.66}  & \cellcolor[HTML]{EFEFEF}82.9           & \multicolumn{1}{c|}{\cellcolor[HTML]{EFEFEF}13.58} & \cellcolor[HTML]{EFEFEF}31.77          & \multicolumn{1}{c|}{\cellcolor[HTML]{EFEFEF}5.29}  & \cellcolor[HTML]{EFEFEF}77.57          \\ 
\multicolumn{1}{|c|}{}                            & ResNet50\_CA                                      & \multicolumn{1}{c|}{28.46}                         & 4.87                                & \multicolumn{1}{c|}{13.95}                         & 18.47                                     & \multicolumn{1}{c|}{28.33}                          & 4.07                                   & \multicolumn{1}{c|}{14.75}                         & 9.83                                   \\ 
\multicolumn{1}{|c|}{}                            & \cellcolor[HTML]{EFEFEF}ResNet50\_CA w/ DropModal & \multicolumn{1}{c|}{\cellcolor[HTML]{EFEFEF}18} & \cellcolor[HTML]{EFEFEF}9.8          & \multicolumn{1}{c|}{\cellcolor[HTML]{EFEFEF}13.72}  & \cellcolor[HTML]{EFEFEF}26.1           & \multicolumn{1}{c|}{\cellcolor[HTML]{EFEFEF}16} & \cellcolor[HTML]{EFEFEF}7.47           & \multicolumn{1}{c|}{\cellcolor[HTML]{EFEFEF}13.21}  & \cellcolor[HTML]{EFEFEF}20.37          \\ \cline{2-10} 
\multicolumn{1}{|c|}{}                            & ViT-Base                                          & \multicolumn{1}{c|}{20.33}                         & 4.07                                   & \multicolumn{1}{c|}{2.5}                           & 84.27                                  & \multicolumn{1}{c|}{29.51}                         & 3.1                                    & \multicolumn{1}{c|}{3.82}                          & 80.87                                  \\ 
\multicolumn{1}{|c|}{}                            & \cellcolor[HTML]{EFEFEF}ViT-Base w/ DropModal     & \multicolumn{1}{c|}{\cellcolor[HTML]{EFEFEF}7.87}  & \cellcolor[HTML]{EFEFEF}40.73          & \multicolumn{1}{c|}{\cellcolor[HTML]{EFEFEF}2.59}  & \cellcolor[HTML]{EFEFEF}80.97          & \multicolumn{1}{c|}{\cellcolor[HTML]{EFEFEF}9.06}  & \cellcolor[HTML]{EFEFEF}30.37          & \multicolumn{1}{c|}{\cellcolor[HTML]{EFEFEF}4.97}  & \cellcolor[HTML]{EFEFEF}79.43          \\ 
\multicolumn{1}{|c|}{}                            & ViT-Base\_SE                                      & \multicolumn{1}{c|}{23.28}                         & 1.5                                    & \multicolumn{1}{c|}{1.87}                          & 92.67                                  & \multicolumn{1}{c|}{37.38}                         & 1.93                                   & \multicolumn{1}{c|}{2.7}                           & 88.2                                   \\ 
\multicolumn{1}{|c|}{}                            & \cellcolor[HTML]{EFEFEF}ViT-Base\_SE w/ DropModal & \multicolumn{1}{c|}{\cellcolor[HTML]{EFEFEF}8.58}  & \cellcolor[HTML]{EFEFEF}36.6           & \multicolumn{1}{c|}{\cellcolor[HTML]{EFEFEF}3.01}  & \cellcolor[HTML]{EFEFEF}77.73          & \multicolumn{1}{c|}{\cellcolor[HTML]{EFEFEF}10.18} & \cellcolor[HTML]{EFEFEF}30.8           & \multicolumn{1}{c|}{\cellcolor[HTML]{EFEFEF}3.03}  & \cellcolor[HTML]{EFEFEF}78.6           \\ 
\multicolumn{1}{|c|}{}                            & ViT-Base\_CA                                      & \multicolumn{1}{c|}{19.4}                          & 13.27                                   & \multicolumn{1}{c|}{1.75}                          & 87.63                                  & \multicolumn{1}{c|}{14.84}                          & 15.07                                 & \multicolumn{1}{c|}{2.43}                          & 78.75                                   \\ 
\multicolumn{1}{|c|}{\multirow{-18}{*}{Unified}}  & \cellcolor[HTML]{EFEFEF}ViT-Base\_CA w/ DropModal & \multicolumn{1}{c|}{\cellcolor[HTML]{EFEFEF}6.04}  & \cellcolor[HTML]{EFEFEF}57.53          & \multicolumn{1}{c|}{\cellcolor[HTML]{EFEFEF}3.85}  & \cellcolor[HTML]{EFEFEF}73.3          & \multicolumn{1}{c|}{\cellcolor[HTML]{EFEFEF}5.97}  & \cellcolor[HTML]{EFEFEF}57.5           & \multicolumn{1}{c|}{\cellcolor[HTML]{EFEFEF}3.91}  & \cellcolor[HTML]{EFEFEF}71.17          \\ \hline

\end{tabular}}
\label{tab:experiment1}
\end{table*}

 \begin{table*}[t]
  \vspace{-0.8em}
\centering
\caption{\textbf{Table 3:} \small{Results of cross-dataset testings on WMCA when trained on CASIA-SURF and CeFA.}}
\vspace{-0.8em}
\resizebox{1.0\textwidth}{!}{\begin{tabular}{|cl|cc|cc|cc|cc|}
\hline
\multicolumn{2}{|c|}{}                                                                                & \multicolumn{2}{c|}{Protocol 1}                                                           & \multicolumn{2}{c|}{Protocol 2}                                                           & \multicolumn{2}{c|}{Protocol 3}                                                           & \multicolumn{2}{c|}{Protocol 4}                                                           \\ \cline{3-10} 
\multicolumn{2}{|c|}{\multirow{-2}{*}{Method}}                                                        & \multicolumn{1}{l|}{ACER(\%) $\downarrow$}                      & \multicolumn{1}{l|}{TPR(\%)@FPR=1\% $\uparrow$} & \multicolumn{1}{l|}{ACER(\%) $\downarrow$}                      & \multicolumn{1}{l|}{TPR(\%)@FPR=1\% $\uparrow$} & \multicolumn{1}{l|}{ACER(\%) $\downarrow$}                      & \multicolumn{1}{l|}{TPR(\%)@FPR=1\% $\uparrow$} & \multicolumn{1}{l|}{ACER(\%) $\downarrow$}                      & \multicolumn{1}{l|}{TPR(\%)@FPR=1\% $\uparrow$} \\ \hline
\multicolumn{1}{|c|}{}                            & CDCN \cite{yu2020searching}                                               & \multicolumn{1}{c|}{34.35}                         & 9.61                                 & \multicolumn{1}{c|}{27.58}                         & 8.17                                 & \multicolumn{1}{c|}{40.62}                         & 3.27                                 & \multicolumn{1}{c|}{28.28}                         & 4.71                                 \\ 
\multicolumn{1}{|c|}{}                            & ResNet50 \cite{he2016deep}                                          & \multicolumn{1}{c|}{36.57}                         & 16.14                                & \multicolumn{1}{c|}{29.98}                         & 12.01                                & \multicolumn{1}{c|}{50}                            & 0.77                                 & \multicolumn{1}{c|}{30.72}                         & 13.74                                \\ 
\multicolumn{1}{|c|}{\multirow{-3}{*}{Separate}} & ViT-Base \cite{dosovitskiy2020image}                                          & \multicolumn{1}{c|}{39.38}                         & 5.96                                 & \multicolumn{1}{c|}{36.81}                         & 11.53                                & \multicolumn{1}{c|}{49.34}                         & 2.88                                 & \multicolumn{1}{c|}{33.08}                         & 3.07                                 \\ \hline
\multicolumn{1}{|c|}{}                            & CDCN                                              & \multicolumn{1}{c|}{50}                            & 7.11                                 & \multicolumn{1}{c|}{27.06}                         & 6.63                                 & \multicolumn{1}{c|}{50}                            & 2.93                                 & \multicolumn{1}{c|}{28.28}                         & 4.71                                 \\ 
\multicolumn{1}{|c|}{}                            & \cellcolor[HTML]{EFEFEF}CDCN w/ DropModal         & \multicolumn{1}{c|}{\cellcolor[HTML]{EFEFEF}32.25} & \cellcolor[HTML]{EFEFEF}7.78         & \multicolumn{1}{c|}{\cellcolor[HTML]{EFEFEF}25.44} & \cellcolor[HTML]{EFEFEF}13.83        & \multicolumn{1}{c|}{\cellcolor[HTML]{EFEFEF}32.95} & \cellcolor[HTML]{EFEFEF}6.24         & \multicolumn{1}{c|}{\cellcolor[HTML]{EFEFEF}26.92} & \cellcolor[HTML]{EFEFEF}11.82        \\ 
\multicolumn{1}{|c|}{}                            & CDCN\_SE                                          & \multicolumn{1}{c|}{43.66}                         & 3.8                                  & \multicolumn{1}{c|}{30.11}                         & 4.8                                  & \multicolumn{1}{c|}{43.3}                          & 2.4                                  & \multicolumn{1}{c|}{28.18}                         & 4.51                                 \\ 
\multicolumn{1}{|c|}{}                            & \cellcolor[HTML]{EFEFEF}CDCN\_SE w/ DropModal     & \multicolumn{1}{c|}{\cellcolor[HTML]{EFEFEF}36.99} & \cellcolor[HTML]{EFEFEF}7.1          & \multicolumn{1}{c|}{\cellcolor[HTML]{EFEFEF}43.55} & \cellcolor[HTML]{EFEFEF}1.73         & \multicolumn{1}{c|}{\cellcolor[HTML]{EFEFEF}42.86} & \cellcolor[HTML]{EFEFEF}4.29         & \multicolumn{1}{c|}{\cellcolor[HTML]{EFEFEF}42.6}  & \cellcolor[HTML]{EFEFEF}2.02         \\ 
\multicolumn{1}{|c|}{}                            & CDCN\_CA                                          & \multicolumn{1}{c|}{40.61}                         & 1.06                                 & \multicolumn{1}{c|}{32.73}                         & 3.5                                  & \multicolumn{1}{c|}{41.26}                         & 1.06                                 & \multicolumn{1}{c|}{40.74}                         & 2.5                                  \\ 
\multicolumn{1}{|c|}{}                            & \cellcolor[HTML]{EFEFEF}CDCN\_CA w/ DropModal     & \multicolumn{1}{c|}{\cellcolor[HTML]{EFEFEF}35.26} & \cellcolor[HTML]{EFEFEF}6.53         & \multicolumn{1}{c|}{\cellcolor[HTML]{EFEFEF}34.49} & \cellcolor[HTML]{EFEFEF}7.2          & \multicolumn{1}{c|}{\cellcolor[HTML]{EFEFEF}35.56} & \cellcolor[HTML]{EFEFEF}7.2          & \multicolumn{1}{c|}{\cellcolor[HTML]{EFEFEF}34.44} & \cellcolor[HTML]{EFEFEF}5.48         \\ \cline{2-10} 
\multicolumn{1}{|c|}{}                            & ResNet50                                          & \multicolumn{1}{c|}{46.02}                         & 4.71                                 & \multicolumn{1}{c|}{35.81}                         & 7.49                                 & \multicolumn{1}{c|}{42.22}                         & 2.21                                 & \multicolumn{1}{c|}{30.72}                          & 13.74                                \\ 
\multicolumn{1}{|c|}{}                            & \cellcolor[HTML]{EFEFEF}ResNet50 w/ DropModal     & \multicolumn{1}{c|}{\cellcolor[HTML]{EFEFEF}46.45} & \cellcolor[HTML]{EFEFEF}11.53        & \multicolumn{1}{c|}{\cellcolor[HTML]{EFEFEF}25.36} & \cellcolor[HTML]{EFEFEF}19.79        & \multicolumn{1}{c|}{\cellcolor[HTML]{EFEFEF}46.53} & \cellcolor[HTML]{EFEFEF}10.28        & \multicolumn{1}{c|}{\cellcolor[HTML]{EFEFEF}19.43} & \cellcolor[HTML]{EFEFEF}19.79        \\ 
\multicolumn{1}{|c|}{}                            & ResNet50\_SE                                      & \multicolumn{1}{c|}{49.52}                         & 1.34                                 & \multicolumn{1}{c|}{28.86}                         & 7.88                                 & \multicolumn{1}{c|}{43.58}                         & 0.67                                 & \multicolumn{1}{c|}{32.55}                         & 7.49                                 \\ 
\multicolumn{1}{|c|}{}                            & \cellcolor[HTML]{EFEFEF}ResNet50\_SE w/ DropModal     & \multicolumn{1}{c|}{\cellcolor[HTML]{EFEFEF}45.31} & \cellcolor[HTML]{EFEFEF}13.26        & \multicolumn{1}{c|}{\cellcolor[HTML]{EFEFEF}30.88} & \cellcolor[HTML]{EFEFEF}13.26        & \multicolumn{1}{c|}{\cellcolor[HTML]{EFEFEF}42.22} & \cellcolor[HTML]{EFEFEF}17.58        & \multicolumn{1}{c|}{\cellcolor[HTML]{EFEFEF}27.41} & \cellcolor[HTML]{EFEFEF}15.95        \\ 
\multicolumn{1}{|c|}{}                            & ResNet50\_CA                                      & \multicolumn{1}{c|}{39.15}                         & 10.66                                 & \multicolumn{1}{c|}{34}                         & 11.24                                 & \multicolumn{1}{c|}{36.9}                         & 8.45                                 & \multicolumn{1}{c|}{34.76}                         & 7.3                                 \\ 
\multicolumn{1}{|c|}{}                            & \cellcolor[HTML]{EFEFEF}ResNet50\_CA w/ DropModal & \multicolumn{1}{c|}{\cellcolor[HTML]{EFEFEF}43.32} & \cellcolor[HTML]{EFEFEF}4.42         & \multicolumn{1}{c|}{\cellcolor[HTML]{EFEFEF}31.94} & \cellcolor[HTML]{EFEFEF}7.3         & \multicolumn{1}{c|}{\cellcolor[HTML]{EFEFEF}47.65} & \cellcolor[HTML]{EFEFEF}5         & \multicolumn{1}{c|}{\cellcolor[HTML]{EFEFEF}37.17} & \cellcolor[HTML]{EFEFEF}8.17         \\ \cline{2-10} 
\multicolumn{1}{|c|}{}                            & ViT-Base                                          & \multicolumn{1}{c|}{48.52}                         & 9.03                                 & \multicolumn{1}{c|}{44.42}                         & 7.59                                 & \multicolumn{1}{c|}{50}                            & 0.38                                 & \multicolumn{1}{c|}{33.08}                         & 3.07                                  \\ 
\multicolumn{1}{|c|}{}                            & \cellcolor[HTML]{EFEFEF}ViT-Base w/ DropModal     & \multicolumn{1}{c|}{\cellcolor[HTML]{EFEFEF}39.06} & \cellcolor[HTML]{EFEFEF}8.26         & \multicolumn{1}{c|}{\cellcolor[HTML]{EFEFEF}30.37} & \cellcolor[HTML]{EFEFEF}15.95        & \multicolumn{1}{c|}{\cellcolor[HTML]{EFEFEF}40.61} & \cellcolor[HTML]{EFEFEF}8.36         & \multicolumn{1}{c|}{\cellcolor[HTML]{EFEFEF}29.51} & \cellcolor[HTML]{EFEFEF}17.2         \\ 
\multicolumn{1}{|c|}{}                            & ViT-Base\_SE                                      & \multicolumn{1}{c|}{49.95}                         & 3.75                                 & \multicolumn{1}{c|}{30.64}                         & 5.09                                 & \multicolumn{1}{c|}{50}                            & 1.83                                 & \multicolumn{1}{c|}{41.03}                         & 5.48                                 \\ 
\multicolumn{1}{|c|}{}                            & \cellcolor[HTML]{EFEFEF}ViT-Base\_SE w/ DropModal & \multicolumn{1}{c|}{\cellcolor[HTML]{EFEFEF}33.48} & \cellcolor[HTML]{EFEFEF}9.51         & \multicolumn{1}{c|}{\cellcolor[HTML]{EFEFEF}30.67} & \cellcolor[HTML]{EFEFEF}9.41         & \multicolumn{1}{c|}{\cellcolor[HTML]{EFEFEF}35.96} & \cellcolor[HTML]{EFEFEF}5.19         & \multicolumn{1}{c|}{\cellcolor[HTML]{EFEFEF}31.33} & \cellcolor[HTML]{EFEFEF}8.55         \\ 
\multicolumn{1}{|c|}{}                            & ViT-Base\_CA                                      & \multicolumn{1}{c|}{35.07}                         & 26.13                               & \multicolumn{1}{c|}{10.07}                          & 50.05                                & \multicolumn{1}{c|}{24.57}                         & 16.04                             & \multicolumn{1}{c|}{20.87}                         & 36.5                               \\ 
\multicolumn{1}{|c|}{\multirow{-18}{*}{Unified}}  & \cellcolor[HTML]{EFEFEF}ViT-Base\_CA w/ DropModal & \multicolumn{1}{c|}{\cellcolor[HTML]{EFEFEF}42.38} & \cellcolor[HTML]{EFEFEF}6.15        & \multicolumn{1}{c|}{\cellcolor[HTML]{EFEFEF}33.87} & \cellcolor[HTML]{EFEFEF}8.17       & \multicolumn{1}{c|}{\cellcolor[HTML]{EFEFEF}37.81} & \cellcolor[HTML]{EFEFEF}6.72        & \multicolumn{1}{c|}{\cellcolor[HTML]{EFEFEF}33.59} & \cellcolor[HTML]{EFEFEF}7.88         \\ \hline

\end{tabular}}
\label{tab:experiment2}
\end{table*}

 \begin{table*}[t]
\centering
\caption{\textbf{Table 4:} \small{Results of `shared' and `unshared' multi-modal settings on intra-dataset testings on CASIA-SURF and CeFA.}}
\vspace{-0.8em}
\resizebox{0.92\textwidth}{!}{\begin{tabular}{|cl|c|c|c|c|cc|}
\hline
\multicolumn{2}{|c|}{\multirow{2}{*}{Method}}                         & Protocol 1        & Protocol 2        & Protocol 3        & Protocol 4        & \multicolumn{2}{c|}{Overall}                  \\ \cline{3-8} 
\multicolumn{2}{|c|}{}                                                & TPR(\%)@FPR=0.1\% $\uparrow$ & TPR(\%)@FPR=0.1\% $\uparrow$ & TPR(\%)@FPR=0.1\% $\uparrow$ & TPR(\%)@FPR=0.1\% $\uparrow$ & \multicolumn{1}{c|}{\#Param. (M)} & \#FLOPs (G) \\ \hline
\multicolumn{1}{|c|}{\multirow{2}{*}{Separate}} & ViT-Base            & 31.33              & 91.27              & 6.5              & 80.87             & \multicolumn{1}{c|}{362.01}      & 132.31     \\
\multicolumn{1}{|c|}{}                          & ViT-Base (unshared) & 43.4              & 91.73             & 1.7               & 88.63              & \multicolumn{1}{c|}{688.58}      & 132.31     \\ \hline
\multicolumn{1}{|c|}{\multirow{2}{*}{Unified}}  & ViT-Base            & 4.07              & 84.27             & 3.1               & 80.87             & \multicolumn{1}{c|}{96.83}       & 199.92     \\
\multicolumn{1}{|c|}{}                          & ViT-Base (unshared) & 0.03              & 89.83             & 0.37              & 88.63             & \multicolumn{1}{c|}{260.12}      & 199.92     \\ \hline

\end{tabular}}
\label{tab:experiment3}
\end{table*}

\subsection{Implementation Details}
The experiments are implemented with Pytorch on one NVIDIA V100 GPU. Three deep models (CDCN~\cite{yu2020searching},  ResNet-50~\cite{he2016deep} and ViT-Base~\cite{dosovitskiy2020image}) are used with batch size 16, 64 and 64, respectively. We use the Adam optimizer with learning rate (lr) of 1e-4 for CDCN and ResNet while AdamW optimizer with lr=1e-5 for ViT-Base. CDCN is trained from scratch with 60 epochs while lr halves in the 30th epoch. Instead of supervision with pseudo depth maps~\cite{yu2020searching}, we follow~\cite{yu2020multi} to supervise CDCN with simple binary maps. In contrast, ResNet50/ViT-Base is finetuned based on the ImageNet/ImageNet-21K pre-trained models with 30 epochs while lr halves in the 20th epoch. 
Direct concatenation is adopted as the default fusion method. For the SE fusion, the intermediate channel numbers are reduced to one eighth of original channels.
The missing modalities are simply blocked as zeros in the testing phase of Protocols 2,3, and 4. To mimic such scenarios in the training phase, similar to~\cite{shen2019facebagnet}, we randomly dropout the \textit{Depth} and \textit{IR} inputs (called DropModal).

\subsection{Intra Testing}
The experimental results of flexible-modal intra-dataset testing on CASIA-SURF and CeFA datasets is shown in Table~\ref{tab:protocol}.
We can see from the first block and first rows in last three blocks that 1) the separated trained ResNet50 and ViT models perform obviously better than the unified counterparts while CDCN performs the opposite; and 2) ViT has higher TPR@FPR=0.1\% than ResNet50 and CDCN on Protocols 2, 3, and 4 with both separated and unified settings, indicating the excellent multi-modal modeling capacities based on global self-attentioned features.

 \vspace{0.2em}
 \noindent\textbf{Impact of fusion modules.} It can be seen from the `Unified' block in Table~\ref{tab:protocol}, compared with directly concatenation fusion, the SE fusion~\cite{casiasurf} has no gains for CDCN, ResNet50, and ViT-Base. In contrast, we can find from the results of `ViT-Base' and `ViT-Base\_CA' that the proposed CA module improves the ViT-Base remarkably (with gains 9.2\%, 3.36\% and 11.93\% TPR@FPR=0.1\% for Protocols 1, 2 and 3, respectively). Despite benefits from CA for ViT-Base backbone, the CA module still generalizes poorly across other architectures (e.g., CDCN and ResNet50). It is still an open question to design architecture-agnostic fusion methods for the flexible-modal FAS benchmark.

 \vspace{0.2em}
\noindent\textbf{Impact of DropModal.} The multi-modal learning is easily dominated by partial-modal features (e.g., \textit{Depth} modality) but neglecting other modalities with relatively weak clues (e.g., \textit{IR} modality). The results of all variants of ResNet50 and ViT-Base on Protocols 1 and 3 are sharply improved with `DropModal', indicating the augmentation with random modality dropout~\cite{shen2019facebagnet} alleviates the modality overfitting issue in intra testings.

\subsection{Cross Testing}
Table~\ref{tab:experiment2} shows the results of flexible-modal cross-dataset testing on WMCA. Due to the domain shifts (e.g., from sensors and sessions) and unseen PAIs, the performance of both separate and unified models are unsatisfactory (ACER$\textgreater$10\%).

 \vspace{0.2em}
\noindent\textbf{Impact of fusion modules.} Similar to the intra-dataset testings, the results with SE fusion~\cite{casiasurf} cannot bring obvious benefits for CDCN, ResNet50, and ViT-Base on cross-dataset testings. As can be seen from the results of `ResNet50\_SE', `ResNet50\_CA', `ViT-Base\_SE', and `ViT-Base\_CA' in Table~\ref{tab:experiment2}, compared with SE fusion, the proposed CA is highly compatible with multi-modal ViT and ResNet50 architectures (especially on Protocols 1, 2, and 3), and improves the cross-dataset testing results dramatically.

 \vspace{0.2em}
\noindent\textbf{Impact of DropModal.} It is reasonable to find in Table~\ref{tab:experiment2} that`DropModal' benefits the cross-testing performance of direct and SE concatenation fusions for all three kinds of models. However, the results of `ResNet50\_CA w/ DropModal' and `ViT-Base\_CA w/ DropModal' indicate that `DropModal' degrades the cross-testing performance for CA fusion for both ResNet50 and ViT-Base backbones. It indicates that training CA with dropped-modality features limit the overall domain generalization capacity.

\subsection{Efficiency Analysis}
Both performance and efficiency are important in flexible-modal benchmark. Here we analyze the efficiency based on ViT-Base with two kinds of settings: separate/unified models and shared/unshared modality branches. As shown on the right part of Table~\ref{tab:experiment3}, compared with separate models, the unified models save more than 50\% parameters but require a bit more FLOPs (due to the fixed tri-modal branch setting in the testing phase) overall 4 protocols. Besides, using unshared backbone for independent modality branch usually brings slight performance improvement but introducing extra huge parameters. Overall, it will be a good tradeoff if the unified models with modality-shared backbones could achieve satisfactory performance.

\section{Conclusion and Future Work} 

In this paper, we establish the first flexible-modal FAS benchmark with both intra- and cross-dataset testings. We also provide sufficient baselines on prevalent deep models and feature fusion strategies for flexible-modal FAS.
However, from the experimental results in Tables~\ref{tab:protocol},~\ref{tab:experiment2} and~\ref{tab:experiment3}, there are still observable performance gaps between separate and unified models in some protocols, as well as the efficiency issue (\#FLOPs).
One limitation and assumption of the benchmark is that it needs all three modalities simultaneously in the training stage but RGB modality usually dominates while partial NIR and Depth modalities are missing under real-world scenarios.

{\small
\bibliographystyle{ieee_fullname}
\bibliography{egbib}
}

\end{document}